\documentclass{article}

\usepackage{PRIMEarxiv}

\usepackage{amsmath}
\usepackage{amsfonts}
\usepackage{amssymb}
\usepackage{amsthm}
\usepackage{algorithm}
\usepackage{algorithmic}
\usepackage{array}
\usepackage{textcomp}
\usepackage{stfloats}
\usepackage[hyphens]{url}
\usepackage{graphicx}
\usepackage{cite}
\usepackage{float}
\usepackage{placeins}
\providecommand{\IEEEmembership}[1]{#1}

\newtheorem{theorem}{Theorem}
\newtheorem{proposition}{Proposition}

\newtheorem{definition}{Definition}

\title{Style as Cover: Deep Image Steganography via Stylized Transmission}

\author{
Qi Li,~\IEEEmembership{Member,~IEEE}\textsuperscript{1},
Jidong Yang\textsuperscript{1,*},\\
Huaike Yu\textsuperscript{1}, Chunpeng Wang,~\IEEEmembership{Senior Member,~IEEE}\textsuperscript{1},
Suo Gao,~\IEEEmembership{Member,~IEEE}\textsuperscript{2},\\
Herbert Ho-Ching Iu,~\IEEEmembership{Fellow,~IEEE}\textsuperscript{3},
Yuantian Miao\textsuperscript{4}, Bin Ma\textsuperscript{1}, and Xiao Chen,~\IEEEmembership{Member,~IEEE}\textsuperscript{4}\\[0.5em]
\parbox{0.96\textwidth}{\normalfont\footnotesize\centering
\textsuperscript{1} Qi Li, Jidong Yang, Huaike Yu, Chunpeng Wang and Bin Ma are with the Key Laboratory of Computing Power Network and Information Security, Ministry of Education, Shandong Computer Science Center, and Shandong Provincial Key Laboratory of Industrial Network and Information System Security, Shandong Fundamental Research Center for Computer Science, Qilu University of Technology (Shandong Academy of Sciences), Jinan 250353, China (e-mail: qluliqi@163.com, jidong\_yang\_paper@163.com, huaikeyu@gmail.com, mpeng1122@163.com, sddxmb@126.com).\\
\textsuperscript{2} Suo Gao is with the School of Information Science and Engineering, Dalian Polytechnic University, Dalian 116034, China (e-mail: gaosuodlpu@163.com).\\
\textsuperscript{3} Herbert Ho-Ching Iu is with the School of Electrical, Electronic and Computer Engineering, The University of Western Australia, Perth, WA 6009, Australia (e-mail: herbert.iu@uwa.edu.au).\\
\textsuperscript{4} Yuantian Miao and Xiao Chen are with the School of Computer and Information Sciences, The University of Newcastle, College of Engineering, Science and Environment, Callaghan, NSW 2308, Australia (e-mail: sky.miao@newcastle.edu.au,xiao.chen@newcastle.edu.au).\\
\textsuperscript{*} Corresponding author: Jidong Yang (e-mail: jidong\_yang\_paper@163.com).
}}

\begin{document}

\maketitle
\begingroup
\renewcommand{\thefootnote}{}
\footnotetext{This work was supported in part by Taishan Scholar under Grant tsqnz20250747; in part by the National Natural Science Foundation under Grant 62502250, Grant 62406051, Grant 62302249, Grant 62541206, and Grant 62272255; in part by the Young Talent of Lifting engineering for Science and Technology in Shandong under Grant SDAST2025QTB030.}
\addtocounter{footnote}{-1}
\endgroup

\begin{abstract}

Image steganography hides secret message within normal images, with most existing works relying on cover-preserving transmission. However, such a paradigm becomes vulnerable once the original cover is exposed or can be reliably approximated. In this paper, we propose StyleStegaNet, a stylized image hiding framework that replaces cover matching with style-concealment transmission. Instead of transmitting a cover-like stego image, StyleStegaNet generates stylized stego images conditioned on publicly available style references, redefining steganography invisibility from cover-preserving concealment to behavior-level camouflage based on style transformation. Such a setting poses a substantial challenge to reliable secret recovery, since neural stylization can significantly alter the feature statistics exploited by deep hiding methods. To address this challenge, StyleStegaNet decouples the overall task into four coordinated stages: stego generation, stylized transmission, structure-preserving reconstruction, and secret recovery. Moreover, StyleStegaNet is optimized with a progressive three-stage training strategy, in which wavelet-domain constraints and perceptual supervision guide the recoverable information toward structural representations. We further provide an analysis showing that secret recoverability is largely restricted to the normalized structural subspace, offering a mechanistic explanation for why directly stylized baselines fail and why a reconstruction-guided recovery path is necessary. Extensive experiments on DIV2K and MS-COCO datasets demonstrate the effectiveness of StyleStegaNet. And few-shot image steganalysis with two deep detectors further shows detection accuracy near random guessing, approximately $51$\%.

\end{abstract}


\section{Introduction}
\label{sec:intro}

Image steganography, as a key technique for information hiding, seeks to achieve secure communication by embedding hidden message into a cover image while preserving both visual fidelity and statistical undetectability. Baluja\cite{baluja2017hiding}\cite{baluja2020hidingwithin} introduced deep steganography as an end-to-end image hiding framework capable of embedding a full-color secret image into a same-resolution cover image. Subsequent studies \cite{jing2021hinet}\cite{lu2021isn}\cite{guan2022deepmih}\cite{xu2022riis} advanced this framework by improving the embedding and decoding mechanisms, resulting in enhanced reconstruction fidelity of the recovered secret image.

Most deep steganography methods rely on the same visual assumption: the stego images transmitted over public channels is required to remain visually close to the original cover images, so that conspicuous appearance discrepancies do not expose the steganographic behavior. In this setting, invisibility is usually treated as a pixel-level fidelity constraint, with the stego image allowed to differ from the cover only by imperceptible perturbations. However, this cover-preserving assumption also exposes a critical security vulnerability. Once the original cover becomes accessible, subtle statistical differences between the cover image and stego image may be enough to reveal the presence of hidden message.

To address this security vulnerability, we replace cover-preserving transmission with stylized transmission, as shown in Figure.~1. Instead of transmitting a stego image that visually resembles the original cover (Figure.~1(a)), our method coverts the stego image into a stylized image suitable for public transmission, where the distributional reshaping induced by stylization can obscure the statistical shifts caused by secret embedding. Since the transmitted image is optimized for stylized appearance instead of cover matching, it breaks the explicit pixel-level correspondence with the original cover, preventing cover leakage from serving as a direct reference. Although stylized stego images provide an effective disguise for covert communication, it also reshapes the secret image reconstruction problem: the receiver must recover the secret image from a stylized image with altered appearance statistics. To solve this problem, we propose StyleStegaNet, a unified architecture that follows the chain, as shown in Figure.~1(b). This is a radical difference from the existing methods~\cite{baluja2017hiding,jing2021hinet,xu2022riis} which are typically formulated as a two-stage process of embedding and extraction for hidden message. Our StyleStegaNet reshapes the transmission paradigm of the steganographic framework and adopts a progressive three-stage training strategy, enabling the concealment of steganographic behavior during public channel while preserving reliable secret recovery. The main contributions of this paper are summarized as follows:

\begin{figure*}[!tbp]
\centering
\includegraphics[width=\textwidth]{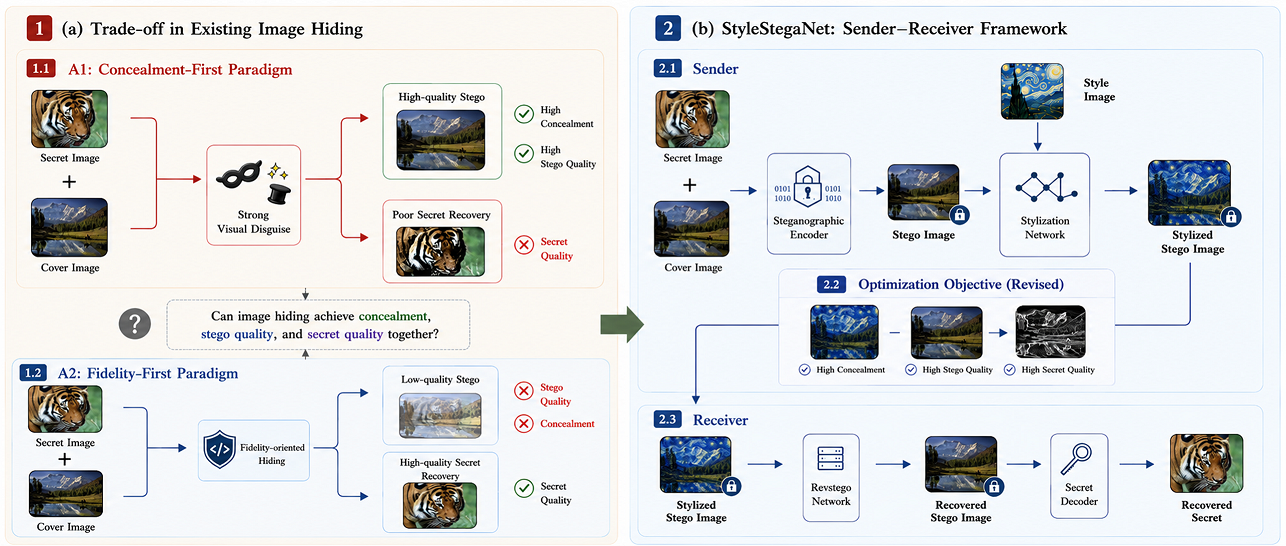}
\caption{Comparison with conventional image hiding. StyleStegaNet transmits a recoverable stylized stego image instead of a stego image close to the cover.}
\label{fig:motivation}
\end{figure*}

(1) We propose StyleStegaNet, a unified sender–receiver framework for stylized full-image hiding. The framework integrates stego generation, frozen stylization,  stego recovery, and secret decoding into a single pipeline, enabling reliable secret extraction from visually normal stylized stego images.

(2) We introduce stylized transmission as a new image steganography paradigm. By transmitting a stylized stego image instead of a cover-preserving stego image, StyleStegaNet redefines steganographic invisibility from pixel-level cover matching to style-based behavioral camouflage, thereby reducing the reference-based security risk caused by cover leakage.

(3) We design a progressive three-stage training strategy with wavelet-domain and perceptual supervision, which stabilizes complex sender–receiver optimization and guides secret-relevant information into stylization-resilient structural representations for reliable secret extraction.

(4) We provide theoretical and empirical validation for StyleStegaNet. The theoretical analysis identifies the recoverable subspace under AdaIN-style stylization and explains the failure of directly stylized baselines. Experiments on DIV2K and MS-COCO  that StyleStegaNet achieves superior secret extraction, stronger steganalysis resistance , and lower secret leakage than existing full-image hiding methods.

The remainder of this paper is organized as follows. In Section II, we give a comprehensive review of related work, focusing on image steganography, neural style transfer and image restoration transformers. In Section III, we explicitly introduce the proposed method, including the network architecture, theoretical analysis and optimization objectives. In Section IV, we describe the experimental setup, followed by a thorough discussion and analysis of the experimental results. In Section V, we give a detailed discussion and limitations. Finally, we conclude this paper and discuss some future directions for this work.

\section{Related Works}
\label{sec:related}

In this section, we review some related works to our proposed methods for stylized image steganography using neural style transfer and image restoration transformers.

\subsection{Image Steganography}
\label{sec:related-hiding}

Traditional steganography hides secret messages by modifying the least significant bits of a cover image or by perturbing coefficients in transform domains such as DCT or DWT. Spatial-domain and frequency-domain methods can tolerate small distortions, but they provide limited capacity and remain vulnerable to modern steganalysis~\cite{pevny2010steganalysis,boroumand2018srnet}.

Deep image hiding extends the payload from bit messages to full images. Baluja~\cite{baluja2017hiding} first showed that a CNN can hide an entire RGB image inside another image. HiDDeN~\cite{zhu2018hidden} introduced a noise layer to improve robustness in message hiding. Weng et al.~\cite{weng2019high} extended this idea to video by using temporal residuals. StegaStamp~\cite{tancik2020stegastamp} embeds hyperlinks into printed photographs and recovers them after physical capture. Distortion-agnostic watermarking~\cite{luo2020distortion} learns to resist a broad range of unknown channel transformations. Invertible neural networks (INNs)~\cite{dinh2015nice,dinh2017realnvp,kingma2018glow} learn bijective mappings between data and latent variables. HiNet~\cite{jing2021hinet} uses affine coupling blocks so that hiding and revealing are implemented as inverse processes in the same network. ISN~\cite{lu2021isn} further increases capacity and shows that a single INN can support large payloads. DeepMIH~\cite{guan2022deepmih} stacks multiple INNs in series and uses an importance map module to hide several secret images in one cover image. RIIS~\cite{xu2022riis} adds a robust recovery path for restoring the secret after JPEG compression, Gaussian noise, or random masking. Xiao et al.~\cite{xiao2020irn} use INNs for invertible image rescaling.

More recent studies further show that deep image steganography has moved from simple encoder--decoder hiding pipelines toward generative, invertible, attention-guided, robust, and multi-image transmission paradigms. Diffusion-based generative steganography has been explored in LDStega~\cite{peng2024ldstega} and Stable-Diffusion-based robust generative hiding~\cite{hu2024stablediffusion}, where the secret is embedded through the latent generative process rather than by merely perturbing an existing cover image. Text-to-image multimodal generative steganography~\cite{jiang2025texttoimage} and guidance-feature-distribution-based generative steganography~\cite{sun2024guidance} further indicate that controllable image generation has become an important carrier construction mechanism. Along the invertible-network line, LiDiNet~\cite{li2024lidinet} reduces the complexity of image-in-image steganography, while iSCMIS~\cite{li2024iscmis} introduces spatial-channel attention for multi-image hiding. Robust image hiding has also been improved by frequency and spatial attention~\cite{zeng2024robust}, color-conversion-based transmission~\cite{li2025color}, and state-space modeling in StegMamba~\cite{luo2025stegmamba}. These recent works confirm the continued importance of balancing payload capacity, visual fidelity, robustness, and steganalysis resistance.

Despite these advances, most existing methods still rely on cover-like, generation-based, or distribution-preserving transmission. The transmitted stego image is usually expected to remain visually close to a cover image, or to follow a natural image distribution generated by a specific model. In contrast, we relax this assumption. In the proposed setting, the transmitted image is a stylized stego image, and the main transformation is deep neural style transfer.

\subsection{Neural Style Transfer}
\label{sec:related-style}

Gatys, Ecker, and Bethge~\cite{gatys2016neural} introduced neural style transfer based on VGG feature matching. AdaIN~\cite{huang2017adain} replaces iterative optimization with a single feedforward pass that aligns per-channel means and variances, and it remains a common baseline for fast stylization. WCT~\cite{li2017wct} uses whitening and coloring transforms for universal style transfer. Chen et al.~\cite{chen2020selfcontained} studied a closer connection between hiding and stylization by treating the stylized image itself as the steganographic host and embedding the content of the input image into it, which allows style transfer to be reversed or applied sequentially. Recent arbitrary style transfer methods have also revisited Transformer-based stylization and contrastive style representation learning to improve style consistency and suppress stylization artifacts~\cite{zhang2024rethink}.

Our formulation uses style transfer in the process of generating stego images, but it does not aim to design a stronger stylization model. We do not embed hidden message into an already stylized image. Instead, we first hide an independent secret image in the cover image to form a stego image, and then apply a fixed stylization process to the stego image. Recoverability then depends on which components of the stego signal survive the style transform, rather than on constructing a standalone reversible container. The goal is covert transmission: the stylized stego image should appear as a coherent stylization while resisting deep steganalysis under a leaked-cover threat model. This differs from reversible stylization methods, which do not target covert full-image payload transmission under a leaked-cover setting.

\subsection{Image Restoration Transformers}
\label{sec:related-restormer}

Recovering a hidden secret from a stylized stego image is closer to image reconstruction than to image generation. Restormer~\cite{zamir2022restormer} introduces a transformer architecture that applies attention along the channel dimension through multi-Dconv head transposed attention and uses a gated-Dconv feedforward network. This design scales to high-resolution inputs with nearly linear computational cost. SwinIR~\cite{liang2021swinir} and Uformer~\cite{wang2022uformer} also use window attention for image restoration. More recently, MambaIR~\cite{guo2024mambair} introduced state-space modeling into image restoration and showed that long-range dependency modeling can also be achieved without relying entirely on self-attention.

This reconstruction perspective is particularly important for stylized image steganography. Conventional deep image hiding methods usually emphasize direct extraction or revealing from a cover-like stego image. In contrast, stylized transmission changes color statistics, texture distributions, and local structural cues before decoding, so direct extraction becomes unreliable when the intermediate carrier is not recovered well. We therefore use a restoration transformer as the recovery backbone of the stego-image reconstruction network, because it can model both global style statistics and local high-frequency details. In our framework, reconstruction is not merely an auxiliary enhancement step; it is the bridge that converts the transmitted stylized stego image back into a representation from which the hidden secret can be reliably extracted.

\section{Method}
\label{sec:method}

\subsection{Overview and Notation}
\label{sec:method-overview}

\begin{figure*}[!tbp]
\centering
\includegraphics[width=\textwidth]{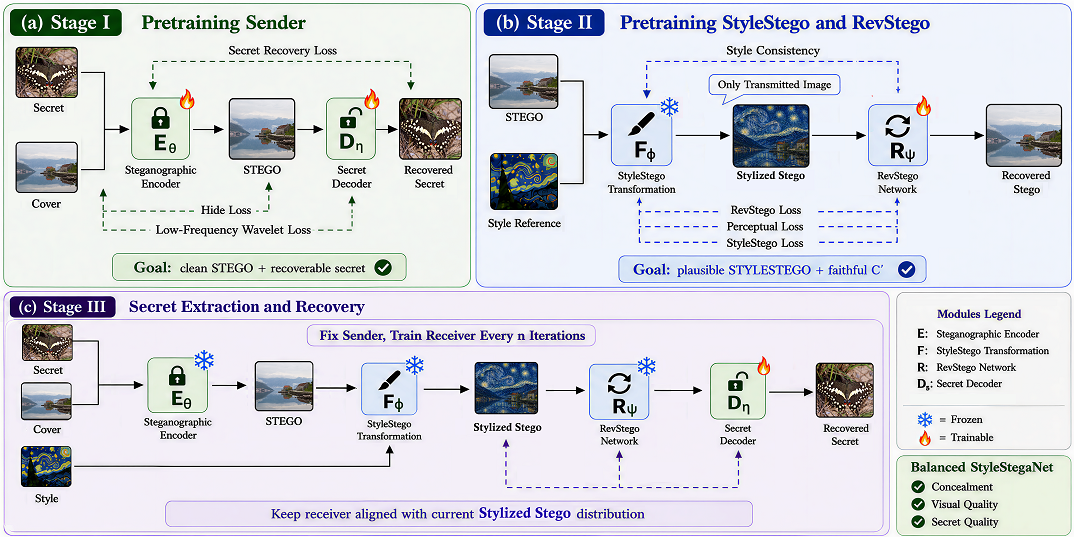}
\refstepcounter{figure}\label{fig:overview}
{\centering\footnotesize Figure.~\thefigure. Three-stage training framework of StyleStegaNet.\par}
\end{figure*}

To ensure clarity and consistency, all the notations used in this paper are summarized in Table~\ref{tab:notation}. After normalization, let $\mathbf{S}\in\mathbb{R}^{H\times W\times 3}$, $\mathbf{C}\in\mathbb{R}^{H\times W\times 3}$, and $\mathbf{Y}\in\mathbb{R}^{H\times W\times 3}$ denote the secret image, cover image, and style reference, respectively. StyleStegaNet follows a sender--receiver architecture. At the sender side, the secret image $\mathbf{S}$ is embedded into the cover image $\mathbf{C}$, resulting in an intermediate stego image $\mathbf{X}$. Guided by the style reference $\mathbf{Y}$, $\mathbf{X}$ is further transformed into a stylized stego image $\mathbf{T}$, which is transmitted over the public channel. At the receiver side, the stego image $\mathbf{X}'$ is recovered from $\mathbf{T}$ by the RevStego network $R_\psi$, and the secret image $\mathbf{S}'$ is subsequently extracted from $\mathbf{X}'$. The framework of our proposed method is shown in Figure~\ref{fig:overview}.

\begin{table}[!t]
\caption{Summary of Notation.}
\label{tab:notation}
\centering
\footnotesize
\setlength{\tabcolsep}{3pt}
\renewcommand{\arraystretch}{1.08}
\begin{tabular}{@{}p{0.24\columnwidth}p{0.68\columnwidth}@{}}
\hline
Notation & Description \\
\hline
$\mathbf{S}$ & Secret image \\
$\mathbf{C}$ & Cover image \\
$\mathbf{Y}$ & Style image \\
$\mathbf{X}$ & Stego image \\
$\mathbf{T}$ & Stylized stego image transmitted by the sender \\
$\mathbf{X}'$ & Recovered stego image \\
$\mathbf{S}'$ & Extracted secret image \\
$E_\theta$ & Steganographic encoder \\
$F_\phi$ & Frozen stylization network \\
$R_\psi$ & RevStego network \\
$D_\eta$ & Secret decoder \\
$\mathcal{H}(\cdot)$ & Haar discrete wavelet transform \\
$\mathrm{LL}(\cdot)$ & Low-frequency Haar sub-band extractor \\
$\mathcal{H}_{\mathrm{HF}}(\cdot)$ & High-frequency Haar sub-band extractor \\
$\hat{\mathbf{x}}$ & Prediction image in reconstruction losses \\
$\mathbf{x}$ & Target image in reconstruction losses \\
\hline
\end{tabular}
\end{table}

In the training phase, each secret-cover pair $(\mathbf{S},\mathbf{C})$ is processed by the Steganographic Encoder $E_\theta$ to produce an intermediate stego image $\mathbf{X}$, which is further converted into a stylized stego image $\mathbf{T}$ with a frozen Stylization Network $F_\phi$. Given $\mathbf{T}$, the receiver-side pipeline sequentially recovers the stego image $\mathbf{X}'$, and extracts the secret image $\mathbf{S}'$ with RevStego network $R_\psi$, and secret decoder $D_\eta$, respectively. That is
\begin{align}
\mathbf{X} &= E_\theta(\mathbf{S},\mathbf{C}), \label{eq:stego}\\
\mathbf{T} &= F_\phi(\mathbf{X},\mathbf{Y}), \label{eq:stylestego}\\
\mathbf{X}' &= R_\psi(\mathbf{T}), \label{eq:revstego}\\
\mathbf{S}' &= D_\eta(\mathbf{X}'), \label{eq:decode-secret}
\end{align}
where only $\mathbf{T}$ is transmitted over the public channel. To make secret reconstruction tractable under stylized transmission, we freeze the stylization network and use it as a fixed transformation module within the overall framework. By stabilizing the stylization process and decoupling the transmitted stego image from cover-like pixel matching, this design mitigates the reference-based detection risk introduced by cover leakage. Note that the style reference $\mathbf{Y}$ and the associated stylization family are treated as public information available to both the sender and the receiver.

\subsection{StyleStego Generation}
\label{sec:method-encoder}

\begin{figure}[!tbp]
\centering
\includegraphics[width=\linewidth]{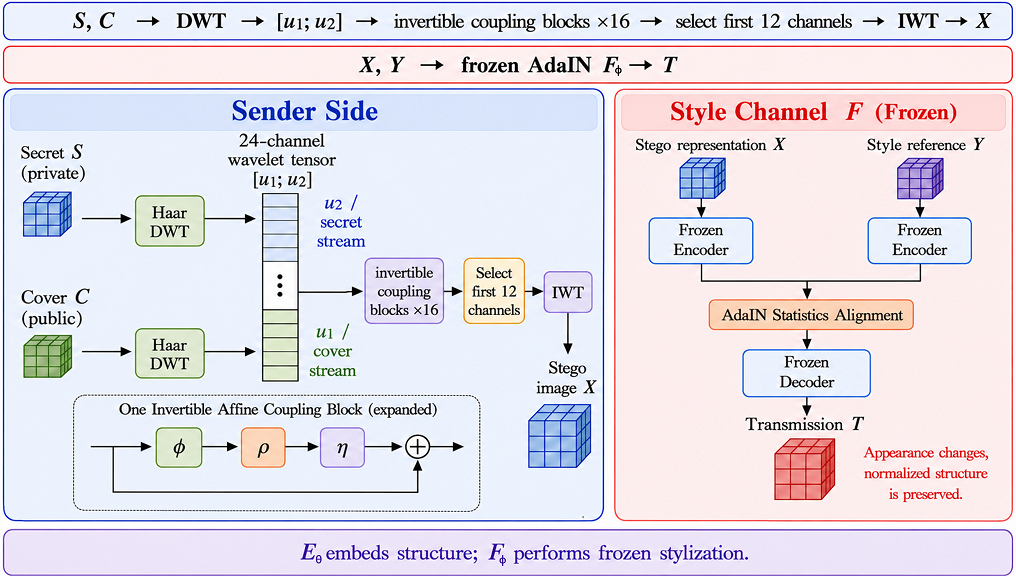}
\caption{StyleStego generation at the sender: a learnable steganographic encoder $E_\theta$ followed by a frozen stylization network $F_\phi$.}
\label{fig:sphm}
\end{figure}

At the sender side, a secret-cover pair is transformed into the transmitted stylized stego image $\mathbf{T}$ through a unified generation stage. This stage consists of a learnable steganographic encoder $E_\theta$, which embeds the secret into an intermediate stego image $\mathbf{X}$, and a frozen stylization network $F_\phi$, which renders $\mathbf{X}$ under the public style reference $\mathbf{Y}$, as shown in Figure~\ref{fig:sphm}. During training, $F_\phi$ is incorporated into the optimization process, while $E_\theta$ is updated to embed secret-relevant information into the high-frequency Haar bands that remain recoverable after stylization. In this way, the stylized operation becomes part of the sender's training objective, not a separate postprocessing step.

\textbf{Steganographic Encoder $E_\theta$:} The Steganographic Encoder $E_\theta$ combines the secret image and the cover image into a stego image $\mathbf{X}$ with an invertible network operating in the Haar wavelet domain. Specifically, Haar transform $\mathcal{H}(\cdot)$ maps each $3$-channel image to a $12$-channel feature representation at half spatial resolution, with one low-frequency (LL) band and three high-frequency bands (LH, HL, HH). The cover and secret wavelet tensors are concatenated into a $24$-channel input and split into a cover stream $u_1$ and a secret stream $u_2$. The two streams are mixed by $16$ invertible affine coupling blocks. Each block applies an additive update $u_1\!\leftarrow\!u_1+\phi(u_2)$ followed by an affine update $u_2\!\leftarrow\!\exp\!\big(2c\,[\sigma(\rho(u_1))-\tfrac12]\big)\odot u_2+\eta(u_1)$, where $\sigma$ is the logistic sigmoid (so the log-scale is bounded by the clamp $c{=}2$), and $\phi,\rho,\eta$ are five-layer residual-dense subnetworks ($3{\times}3$ convolutions, growth width $32$, LeakyReLU, last layer zero-initialized). An inverse Haar transform maps the cover stream back to the Stego image $\mathbf{X}$. Because the network is bijective, hiding and revealing share the same invertible codec weights; the reverse pass that reconstructs the secret is used by the decoder (Section~\ref{sec:method-decoder}). The low-frequency supervision in Section~\ref{sec:method-loss} keeps the LL band close to the cover, so the payload is routed into the high-frequency Haar bands. 

\textbf{Stylization Network $F_\phi$:} The frozen stylization network $F_\phi$ renders the stego image $\mathbf{X}$ as the transmitted stylized stego image $\mathbf{T}$ under the public style reference $\mathbf{Y}$, as shown in Figure.~\ref{fig:sphm}. We implement it with a frozen AdaIN transform~\cite{huang2017adain}, built from relu4\_1 in VGG19 and a pretrained decoder. It aligns the channel-wise mean and standard deviation of the content feature with those of the style feature:
\begin{equation}
\mathrm{AdaIN}(\mathbf{c},\mathbf{y})=\sigma(\mathbf{y})\cdot\frac{\mathbf{c}-\mu(\mathbf{c})}{\sigma(\mathbf{c})}+\mu(\mathbf{y}),
\label{eq:adain-op}
\end{equation}
where $\mu$ and $\sigma$ are computed per channel over spatial locations. Here, stylization is part of the learned sender rather than a postprocessing step attached to an existing hiding encoder: The transmitted stylized stego image $\mathbf{T}$ is optimized to exhibit a natural stylized appearance while preserving sufficient recoverable information for receiver.

The primary focus of this paper is to concretely demonstrate that embedded secret image can be reliably reconstructed from a stylized stego image at the receiver side. Therefore, the AdaIN stylization module is critical to our method as it determines which components of the stego image can remain recoverable after stylized transformation and consequently, where the encoder should allocate secret information.

\begin{theorem}[Affine null space of AdaIN]
\label{prop:adain-null}
Let $\mathbf{c}$ be a content feature map, and let $a>0$ and $b\in\mathbb{R}$ be per-channel constants. Then, for any style feature map $\mathbf{y}$,
\begin{equation}
\mathrm{AdaIN}(a\mathbf{c}+b,\,\mathbf{y})=\mathrm{AdaIN}(\mathbf{c},\mathbf{y}).
\label{eq:adain-invariance}
\end{equation}
Hence $\mathrm{AdaIN}(\cdot,\mathbf{y})$ depends on the content only through the per-channel normalized feature $\hat{\mathbf{c}}=(\mathbf{c}-\mu(\mathbf{c}))/\sigma(\mathbf{c})$, and remains constant over the per-channel affine family of first and second moment reparametrizations.
\end{theorem}

For a per-channel affine map with $a>0$, $\mu(a\mathbf{c}+b)=a\mu(\mathbf{c})+b$ and $\sigma(a\mathbf{c}+b)=a\sigma(\mathbf{c})$. Substituting these identities into \eqref{eq:adain-op}, the normalized term becomes
\[
\frac{(a\mathbf{c}+b)-(a\mu(\mathbf{c})+b)}{a\sigma(\mathbf{c})}=\frac{\mathbf{c}-\mu(\mathbf{c})}{\sigma(\mathbf{c})}=\hat{\mathbf{c}},
\]
so the output is $\sigma(\mathbf{y})\,\hat{\mathbf{c}}+\mu(\mathbf{y})$, which does not depend on $(a,b)$.

A content feature can be decomposed into its per-channel statistics $(\mu(\mathbf{c}),\sigma(\mathbf{c}))$ and its normalized structure $\hat{\mathbf{c}}$. Theorem~\ref{prop:adain-null} shows that the statistic component lies in the null space of the transform, whereas $\hat{\mathbf{c}}$ is preserved through a known affine mapping. This result identifies the subspace through which a recoverable payload can be transmitted.

\begin{theorem}[Subspace that survives stylization]
\label{cor:survivable}
Let $\hat{\mathbf{c}}_{\mathbf{X}}$ denote the normalized feature of the stego image $\mathbf{X}$ under the frozen stylization encoder. The information channel induced by $F_\phi$ has zero capacity on the statistic component and preserves the structure component $\hat{\mathbf{c}}_{\mathbf{X}}$. Consequently, any secret information that the receiver can recover from $\mathbf{T}$ is measurable with respect to $\hat{\mathbf{c}}_{\mathbf{X}}$:
\begin{equation}
I(\mathbf{S};\mathbf{T})\le I\!\left(\mathbf{S};\hat{\mathbf{c}}_{\mathbf{X}}\right).
\label{eq:survivable-mi}
\end{equation}
\end{theorem}

The invariance is exact in the feature space. Because the stylization encoder and decoder are fixed and approximately invertible on natural image features, survivability at the pixel level follows the behavior at the feature level in practice. The recovery network $R_\psi$ therefore learns an inverse mapping on the preserved subspace instead of an analytic inverse of $F_\phi$. We do not claim lossless pixel recovery. Instead, we claim that the recoverable component is confined to $\hat{\mathbf{c}}_{\mathbf{X}}$. The conditioning of this inverse is determined by the style scales.

\begin{proposition}[Recovery conditioning on the structure subspace]
\label{prop:conditioning}
On the structure subspace, the frozen channel acts as the per-channel affine map
$\hat{\mathbf{c}}\mapsto\sigma(\mathbf{Y})\odot\hat{\mathbf{c}}+\mu(\mathbf{Y})$,
whose Jacobian is $\mathrm{diag}(\sigma(\mathbf{Y}))$. It is therefore
bi-Lipschitz with constants $\ell=\min_c\sigma_c(\mathbf{Y})$ and
$L=\max_c\sigma_c(\mathbf{Y})$. The optimal structure-domain inverse satisfies
\begin{equation}
\big\|\hat{\mathbf{c}}_{\mathbf{X}'}-\hat{\mathbf{c}}_{\mathbf{X}}\big\|
\le
\frac{1}{\ell}
\big\|\mathbf{T}-F_\phi(\mathbf{X},\mathbf{Y})\big\|,
\label{eq:conditioning}
\end{equation}
where the condition number is
\begin{equation}
\kappa_F
=
\frac{L}{\ell}
=
\frac{\max_c\sigma_c(\mathbf{Y})}{\min_c\sigma_c(\mathbf{Y})}.
\label{eq:conditioning-kappa}
\end{equation}
\end{proposition}

By Theorem~\ref{prop:adain-null}, $F_\phi$ depends on the stego only through $\hat{\mathbf{c}}_{\mathbf{X}}$, on which it is the affine map above. The singular values of $\mathrm{diag}(\sigma(\mathbf{Y}))$ are the per-channel standard deviations $\sigma_c(\mathbf{Y})$, which give the bi-Lipschitz constants; inverting the affine map yields \eqref{eq:conditioning}.

Proposition~\ref{prop:conditioning} replaces the informal notion of approximate invertibility: $R_\psi$ can invert $F_\phi$ on the structure subspace because that channel is affine and well-conditioned whenever the per-channel style scales $\sigma_c(\mathbf{Y})$ are bounded away from zero.

Theorem~\ref{cor:survivable} provides one explanation for why payloads tied to feature statistics are vulnerable to AdaIN-style transmission, and it motivates the design of StyleStegaNet. A hiding scheme whose payload changes channel-level energy or global statistics, including LSB and most spatial or transform domain methods, is suppressed by \eqref{eq:adain-op}. This suppression accounts for the $7$--$11$~dB collapse of stylized baselines in Table~\ref{tab:sota-style}. In contrast, the low-frequency loss \eqref{eq:Llf} and the Haar high-frequency loss \eqref{eq:Lhf} encourage the secret information to occupy the normalized, spatially high-frequency structure preserved by AdaIN. Thus, the frequency-domain objective is consistent with the feature-statistic view: both encourage the payload to reside in components less affected by AdaIN-style normalization.

\subsection{Secret Image Extraction}
\label{sec:method-revstego}
\label{sec:method-decoder}

\begin{figure}[!tbp]
\centering
\includegraphics[width=\linewidth]{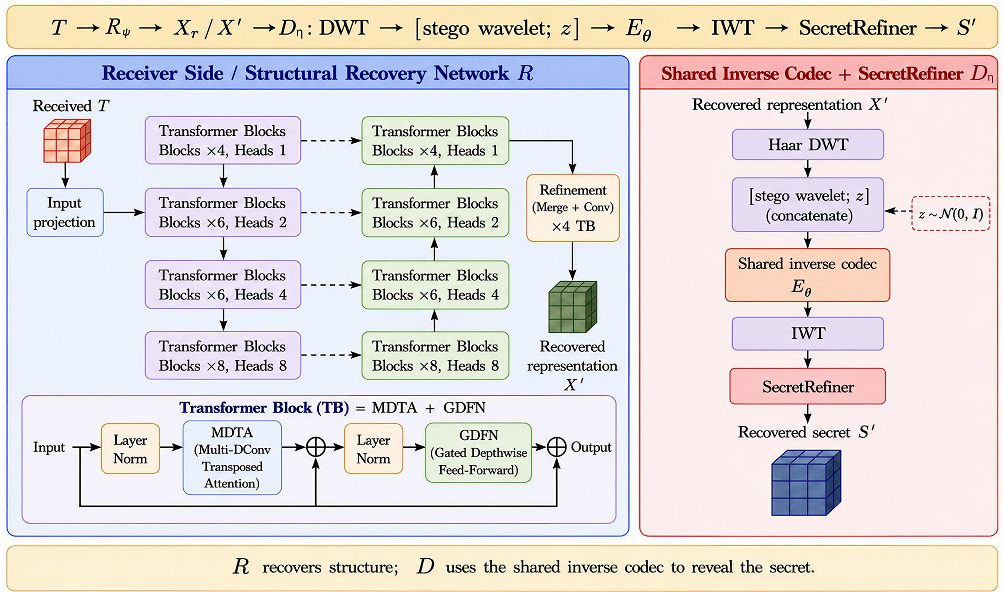}
\caption{Architecture of the RevStego network $R_\psi$ and secret decoder $D_\eta$.}
\label{fig:rsrn}
\end{figure}

From the perspective of the receiver side, recoverable information in the stylized stego image must be first mapped by the RevStego Network $R_\psi$ into a stego image representation $\mathbf{X}'$, from which the Secret Decoder $D_\eta$ subsequently extracts the secret image, as shown in Figure.~\ref{fig:rsrn}. Therefore, $R_\psi$ serves as the critical module in the secret recovery path after stylized transmission, and its ability to recover stego structures and secret information directly determines the completeness and reconstruction quality of the extracted secret image.

\textbf{RevStego Network $R_\psi$:} Given the transmitted stylized stego image $\mathbf{T}$, the receiver first reconstructs a structural stego representation $\mathbf{X}'$ with a four-scale encoder--decoder Transformer. The input is projected to a feature width of $48$ and processed across four scales, where the encoder--decoder stages contain $(4,6,6,8)$ Transformer blocks with $(1,2,4,8)$ attention heads, respectively. Skip connections are employed to preserve structural information, and four refinement blocks are appended at full resolution to enhance the reconstructed details. Following Restormer~\cite{zamir2022restormer}, each block consists of a Multi-DConv Transposed Attention (MDTA) module for efficient global context modeling and a Gated Depthwise Feed-forward Network (GDFN) with an expansion factor of $2.66$ for local detail restoration. Reflection padding is applied so that the input resolution is divisible by $8$. The recovered feature representation is then projected into a reconstructed stego image $\mathbf{X}'$, which serves as the input to the secret decoder.

\textbf{Secret Decoder $D_\eta$:} The secret decoder $D_\eta$ reconstructs the secret image from the reconstructed stego image $\mathbf{X}'$ in two steps. It first performs the shared invertible codec in reverse. Specifically, the steganographic encoder $E_\theta$ is applied in the inverse direction with shared parameters, while the secret information is replaced by a Gaussian latent variable $\mathbf{z}$. The resulting coarse secret estimate is then passed to a lightweight restoration module for detail refinement. The refiner adopts a compact Restormer backbone~\cite{zamir2022restormer} with width $36$, $(2,3,3,4)$ blocks, $(1,2,4,8)$ attention heads, and two refinement blocks at full resolution. The refiner is initialized to approximate an identity mapping, ensuring that early decoding remains anchored to the inverse codec. The whole decoder is trained under the supervision of the ground-truth secret image $\mathbf{S}$, with the high-frequency loss introduced in Section~\ref{sec:method-loss} further preserving texture details in the reconstructed secret image $\mathbf{S}'$.

\subsection{Training Objective}
\label{sec:method-loss}

StyleStegaNet is optimized using pixel, perceptual, and wavelet loss functions. We use $\|\cdot\|_1$ to denote the pixel-wise $L_1$ distance and LPIPS~\cite{zhang2018lpips} as the perceptual metric.

 The hide loss is
\begin{equation}
\mathcal{L}_{\mathrm{hide}}=\|\mathbf{X}-\mathbf{C}\|_1. \label{eq:Lhide}
\end{equation}

 The low-frequency wavelet loss~\cite{jing2021hinet} encourages secret information to be placed in high-frequency Haar bands, which improves resistance to steganalysis:
\begin{equation}
\mathcal{L}_{\mathrm{lf}}=\|\mathrm{LL}(\mathbf{X})-\mathrm{LL}(\mathbf{C})\|_1,
\label{eq:Llf}
\end{equation}
where $\mathrm{LL}(\cdot)$ extracts the low-frequency sub-band of the Haar DWT.

 The Haar high-frequency loss aligns the HL, LH, and HH sub-bands of a prediction with those of the corresponding target:
\begin{equation}
\mathcal{L}_{\mathrm{hf}}(\hat{\mathbf{x}},\mathbf{x})=\big\|\mathcal{H}_{\mathrm{HF}}(\hat{\mathbf{x}})-\mathcal{H}_{\mathrm{HF}}(\mathbf{x})\big\|_1.
\label{eq:Lhf}
\end{equation}
where $\hat{\mathbf{x}}$ and $\mathbf{x}$ denote the prediction image and target image defined in Table~\ref{tab:notation}, respectively, and $\mathcal{H}_{\mathrm{HF}}(\cdot)$ extracts the concatenated high-frequency Haar sub-bands (LH, HL, HH). This generic loss closes the receiver-side training loop by supervising $\mathbf{X}'$ against $\mathbf{X}$ after RevStego recovery and $\mathbf{S}'$ against $\mathbf{S}$ after secret decoding.

 Stego recovery loss for $\mathbf{X}'$:
\begin{equation}
\mathcal{L}_{\mathrm{rec}}=\|\mathbf{X}'-\mathbf{X}\|_1+\lambda_p\,\mathrm{LPIPS}(\mathbf{X}',\mathbf{X}),
\label{eq:Lrecon}
\end{equation}
with $\lambda_p{=}0.1$ in our experiments. The secret recovery loss for $\mathbf{S}'$ is
\begin{equation}
\mathcal{L}_{\mathrm{sec}}
=
\|\mathbf{S}'-\mathbf{S}\|_1
+
\lambda_p\,\mathrm{LPIPS}(\mathbf{S}',\mathbf{S}).
\label{eq:Lrev}
\end{equation}

In our experiments, StyleStegaNet is trained in three stages with weighted objectives. In Stage 1, the stego generation module is pretrained to learn a stable embedding-and-extraction initialization before stylized transmission is introduced:
\begin{equation}
\mathcal{L}_{\mathrm{S1}}
=
\mathcal{L}_{\mathrm{hide}}
+
5\,\mathcal{L}_{\mathrm{sec}}
+
\mathcal{L}_{\mathrm{lf}},
\label{eq:stage1}
\end{equation}
where the coefficient $5$ emphasizes secret recovery during initialization. In Stage 2, the RevStego network is trained to recover the stego representation from the stylized stego image:
\begin{equation}
\mathcal{L}_{\mathrm{S2}}
=
\mathcal{L}_{\mathrm{rec}}
+
0.2\,\mathcal{L}_{\mathrm{hf}}(\mathbf{X}',\mathbf{X}),
\label{eq:stage2}
\end{equation}
where the coefficient $0.2$ controls the high-frequency stego-recovery term. In Stage 3, all learnable networks are jointly optimized in an end-to-end manner to align stylized transmission, stego reconstruction, and secret decoding:
\begin{equation}
\mathcal{L}_{\mathrm{S3}}
=
\mathcal{L}_{\mathrm{S2}}
+
\mathcal{L}_{\mathrm{sec}}
+
0.5\,\mathcal{L}_{\mathrm{hf}}(\mathbf{S}',\mathbf{S}),
\label{eq:stage3}
\end{equation}
where the coefficient $0.5$ controls the high-frequency secret-recovery term.

\subsection{Progressive Three-Stage Training}
\label{sec:method-training}

Direct end-to-end training from random initialization is prone to unstable optimization, as the sender and receiver sides differ in parameter scale, optimization objectives, and supervisory requirements. Therefore, we train StyleStegaNet with a progressive three-stage strategy, where each stage is warm-started from the checkpoint of the previous stage.

 Stage 1: We pretrain the steganographic encoder for $35$K iterations with $\mathcal{L}_{\mathrm{S1}}$, establishing a stable embedding-and-extraction initialization in which the stego image $\mathbf{X}$ remains close to the cover image $\mathbf{C}$ while retaining recoverable secret information. In this stage, the steganographic encoder is considered ready for the second-stage training once the PSNR of the stego image exceeds $38$~dB and the PSNR of the extracted secret image exceeds $35$~dB, indicating that it has learned a stable embedding-and-extraction capability.

 Stage 2: The RevStego network is trained for $80$K iterations with $\mathcal{L}_{\mathrm{S2}}$, while the stylization network is kept frozen. In this stage, it is considered acceptable when the reconstructed stego image $\mathbf{X}'$ achieves a PSNR above $32$~dB, and the stylized stego image $\mathbf{T}$ remains stylistically aligned with the style reference $\mathbf{Y}$ while exhibiting a natural visual appearance.

 Stage 3: The steganographic encoder is frozen, while the RevStego network and the secret decoder are jointly trained for $30$K iterations with $\mathcal{L}_{\mathrm{S3}}$ in \eqref{eq:stage3}. To preserve the reconstruction details learned in the previous stage, a smaller learning rate is used for the reconstruction path to improve secret extraction without destabilizing the learned post-stylization reconstruction mapping. This stage is considered acceptable when the reconstructed secret image $\mathbf{S}'$ reaches a PSNR of about $28$~dB.

We optimize the model with AdamW~\cite{loshchilov2019adamw}, using a base learning rate of $10^{-4}$ and a weight decay of $10^{-5}$. Training is performed on a single GPU with bfloat16 mixed precision. CUDA OOM events trigger an automatic fallback in batch size, for example $8{\to}4{\to}2{\to}1$ in Stages~2--3.

\FloatBarrier

\subsection{Theoretical Analysis: Why Stylized Hiding Works}
\label{sec:method-theory}

In the leaked-cover setting, detectability becomes a test between a secret-bearing stylized stego image and a null stylization generated from the same cover and style. This test connects the survivable structure component in Theorem~\ref{cor:survivable} to the observable gap between transmitted image distributions.

We consider the strongest passive adversary used in our diagnostics, as shown in Figure~\ref{fig:residual-comparison}. The detector knows the cover $\mathbf{C}$, or an accurate estimate of it, the public style image $\mathbf{Y}$, and the stylization family. Its goal is to decide whether a transmitted image carries a secret. The task of the adversary is not to recover $\mathbf{S}$, since the legitimate receiver is designed to do so, but to detect the presence of a hidden secret. This setting gives the following balanced hypothesis test:
\begin{equation}
\begin{aligned}
H_0:\;& \mathbf{T}\sim P_0 := \text{law of } F_\phi\!\big(E_\theta(\mathbf{S}_0,\mathbf{C}),\mathbf{Y}\big),\\
H_1:\;& \mathbf{T}\sim P_1 := \text{law of } F_\phi\!\big(E_\theta(\mathbf{S},\mathbf{C}),\mathbf{Y}\big),
\end{aligned}
\label{eq:hypotheses}
\end{equation}
conditioned on $(\mathbf{C},\mathbf{Y})$, where $\mathbf{S}_0$ denotes the null embedding without a secret. Both $P_0$ and $P_1$ are stylizations of the same cover under the same public style. As a result, the residual $\mathbf{T}-\mathbf{C}$ is dominated by a public transform that is independent of the secret, while the secret affects only the $\hat{\mathbf{c}}_{\mathbf{X}}$ component in \eqref{eq:survivable-mi}.

\begin{definition}[Cachin security relative to the style manifold]
\label{def:cachin}
A stylized hider is $\varepsilon$-secure under the leaked-cover model if $D_{\mathrm{KL}}(P_1\,\|\,P_0)\le\varepsilon$, i.e.\ the transmitted law is $\varepsilon$-close in KL to the legitimate secret-free stylization of the same cover and style.
\end{definition}

\begin{proposition}[Detectability bound]
\label{prop:detect}
For the balanced cover-known test \eqref{eq:hypotheses}, the accuracy of any detector satisfies
\begin{equation}
\mathrm{Acc}\;\le\;\tfrac12+\tfrac12\,\delta_{\mathrm{TV}}(P_0,P_1)\;\le\;\tfrac12+\sqrt{\varepsilon/8}.
\label{eq:detect-bound}
\end{equation}
\end{proposition}

The minimum error probability of the optimal balanced test is $\tfrac12\big(1-\delta_{\mathrm{TV}}(P_0,P_1)\big)$, so the accuracy is at most $\tfrac12\big(1+\delta_{\mathrm{TV}}\big)$. Pinsker's inequality gives $\delta_{\mathrm{TV}}\le\sqrt{\tfrac12 D_{\mathrm{KL}}(P_1\|P_0)}\le\sqrt{\varepsilon/2}$. Substitution yields $\tfrac12+\tfrac12\sqrt{\varepsilon/2}=\tfrac12+\sqrt{\varepsilon/8}$.

Under Proposition~\ref{prop:detect}, a detector can exceed $50\%$ accuracy only to the extent that $P_1$ separates from the secret-free stylization law $P_0$ in KL.

\begin{proposition}[Embedding-bounded detectability]
\label{prop:embed-detect}
Under a Gaussian feature model in which $P_0$ and $P_1$ share a common feature covariance $\Sigma$, the gap of Definition~\ref{def:cachin} satisfies
\begin{equation}
\begin{aligned}
\varepsilon
&=D_{\mathrm{KL}}(P_1\,\|\,P_0)\\
&\le \kappa\,\mathbb{E}\big\|\hat{\mathbf{c}}_{\mathbf{X}}-\hat{\mathbf{c}}_{\mathbf{X}_0}\big\|^2,\\
\kappa
&=\tfrac12\big\|\Sigma^{-1/2}\mathrm{diag}(\sigma(\mathbf{Y}))\big\|_{\mathrm{op}}^2,
\end{aligned}
\label{eq:embed-kl}
\end{equation}
where $\hat{\mathbf{c}}_{\mathbf{X}}$ and $\hat{\mathbf{c}}_{\mathbf{X}_0}$ are the normalized structure features of the stego image containing the real secret and the corresponding null stego image. Combined with \eqref{eq:detect-bound},
\begin{equation}
\mathrm{Acc}\;\le\;\tfrac12+\sqrt{\kappa/8}\;\,\mathrm{RMS}\big(\hat{\mathbf{c}}_{\mathbf{X}}-\hat{\mathbf{c}}_{\mathbf{X}_0}\big).
\label{eq:embed-acc}
\end{equation}
\end{proposition}

By Theorem~\ref{prop:adain-null}, in feature space $\mathbf{T}=\mathrm{diag}(\sigma(\mathbf{Y}))\,\hat{\mathbf{c}}_{\mathbf{X}}+\mu(\mathbf{Y})$, so $P_0$ and $P_1$ are Gaussians with equal covariance $\Sigma$ and mean gap $\mathrm{diag}(\sigma(\mathbf{Y}))(\hat{\mathbf{c}}_{\mathbf{X}}-\hat{\mathbf{c}}_{\mathbf{X}_0})$. The Gaussian KL equals $\tfrac12\big\|\Sigma^{-1/2}\mathrm{diag}(\sigma(\mathbf{Y}))(\hat{\mathbf{c}}_{\mathbf{X}}-\hat{\mathbf{c}}_{\mathbf{X}_0})\big\|^2\le\kappa\,\|\hat{\mathbf{c}}_{\mathbf{X}}-\hat{\mathbf{c}}_{\mathbf{X}_0}\|^2$. Taking expectations and applying \eqref{eq:detect-bound} with the monotonicity of the square root gives \eqref{eq:embed-acc}.

Proposition~\ref{prop:embed-detect} shows that the reduced detectability mainly comes from applying the same frozen stylization process to both hypotheses, without explicitly optimizing a separate style loss. Because both hypotheses pass through the same frozen stylization of the same cover and style (Theorem~\ref{prop:adain-null}), the only signal induced by the secret lies in the surviving structure, and the hide loss \eqref{eq:Lhide} together with the low-frequency loss \eqref{eq:Llf} keeps the structure perturbation $\hat{\mathbf{c}}_{\mathbf{X}}-\hat{\mathbf{c}}_{\mathbf{X}_0}$ small. This is consistent with the measured accuracy near $51\%$ instead of above $90\%$.

\paragraph{The cost of undetectability.}
Security has an inherent cost: a transmitted image that carries recoverable secret information cannot lie exactly on the secret-free stylization manifold.

\begin{proposition}[Security--recoverability tradeoff]
\label{prop:tradeoff}
Let the secret-recovery distortion be $\Delta=\mathbb{E}\,\|\mathbf{S}'-\mathbf{S}\|^2$ and define the security--distortion function
\begin{equation}
\mathcal{E}(\Delta)=\min_{E_\theta:\;\mathbb{E}\|\mathbf{S}'-\mathbf{S}\|^2\le\Delta}\; D_{\mathrm{KL}}(P_1\,\|\,P_0).
\label{eq:sec-dist}
\end{equation}
Then (i) the recoverable secret information is bounded by the channel, $I(\mathbf{S};\mathbf{S}')\le I(\mathbf{S};\mathbf{T})$; and (ii) $\mathcal{E}(\Delta)$ is non-increasing and convex in $\Delta$.
\end{proposition}

(i) The pipeline forms the Markov chain $\mathbf{S}\!\to\!\mathbf{T}\!\to\!\mathbf{S}'$, because the receiver acts only on $\mathbf{T}$. The data-processing inequality therefore gives $I(\mathbf{S};\mathbf{S}')\le I(\mathbf{S};\mathbf{T})$. (ii) Increasing the distortion budget $\Delta$ relaxes the feasible set in \eqref{eq:sec-dist}, so the minimum cannot increase, which proves monotonicity. For convexity, take feasible senders for budgets $\Delta_1$ and $\Delta_2$, and time-share them with probability $t$. The resulting mixture satisfies the budget $t\Delta_1+(1-t)\Delta_2$. Since $D_{\mathrm{KL}}(\cdot\,\|\,P_0)$ is convex in its first argument, the mixed transmitted law attains at most $t\,\mathcal{E}(\Delta_1)+(1-t)\,\mathcal{E}(\Delta_2)$.

Proposition~\ref{prop:tradeoff} gives an interpretation of the joint objective $\mathcal{L}_{\mathrm{S3}}$ in \eqref{eq:stage3}. The objective is a Lagrangian relaxation of minimizing detectability under a recovery budget, and the coefficients in the staged objective choose one operating point on the frontier $\mathcal{E}(\Delta)$. This frontier differs from the training-time effects isolated by the ablation in Table~\ref{tab:ablation}, which identify the components that keep the inverse problem well-conditioned in practice. The frontier states that security and recoverability are in tension in principle, while the ablation shows that recovery depends on which signal components survive the style transform, not on surface appearance.

\FloatBarrier

\section{Experimental Results}
\label{sec:experiments}

\subsection{Experimental Settings}
\label{sec:exp-setting}

\textbf{Dataset.} We train StyleStegaNet on DIV2K~\cite{agustsson2017div2k} using $2{,}000$ fixed cover--secret pairs cropped to $256{\times}256$, where covers and secrets are independently sampled. Style references are sampled from the WikiArt dataset~\cite{wikiart2015}, which contains only painting images. For evaluation, we construct $200$ fixed DIV2K pairs and $200$ fixed MS-COCO pairs~\cite{lin2014coco}, with all test images resized to $512{\times}512$. Each test pair is assigned a deterministic style reference from the WikiArt dataset, ensuring that the cover--secret--style triplets remain identical across all compared methods. StyleStegaNet and all baselines are evaluated under the same training and testing settings, including identical data splits, image resolutions, and style references.

\textbf{Implementation Details.} We use AdamW with a learning rate of $10^{-4}$ and a weight decay of $10^{-5}$, and apply a learning rate of $4{\times}10^{-5}$ to the recovery path in Stage~3. All experiments are performed on a single NVIDIA H20 GPU. In Stages~2 and~3, we employ a memory-adaptive batch-size strategy to handle GPU memory limitations, progressively reducing the batch size from $8$ to $4$, $2$, and $1$ when necessary.

\textbf{Evaluation Metrics.} StyleStegaNet was quantitatively evaluated using Peak Signal-to-Noise Ratio (PSNR), Structural Similarity Index (SSIM)~\cite{wang2004ssim}, average pixel difference (APD), and root mean square error (RMSE), where higher PSNR and SSIM and lower APD and RMSE indicate better reconstruction and extraction quality. Unless otherwise stated, the tables report PSNR, SSIM, APD, and RMSE. We evaluate four image pairs. Stego fidelity is measured on $(\mathbf{X},\mathbf{C})$ and quantifies the difference between the stego image and the original cover before stylization. Stylized stego image quality is assessed by the perceptual style consistency between $\mathbf{T}$ and $\mathbf{Y}$, together with human inspection of the transmitted image. Stego recovery is measured on $(\mathbf{X}',\mathbf{X})$ and evaluates how well the receiver reconstructs the stego image from the transmitted stylized stego image. Secret recovery is measured on $(\mathbf{S}',\mathbf{S})$ and evaluates the accuracy of the reconstructed secret.

 Steganalysis protocol. We further evaluate detectability with SRNet~\cite{boroumand2018srnet} and SiaStegNet~\cite{you2021siastegnet}. For each hiding method, detector, and training size $N\in\{10,20,\ldots,100\}$, we train a separate binary classifier on $N$ cover--stego pairs and test it on $200$ held-out pairs, corresponding to $400$ test images. We report the test accuracy at the selected checkpoint. The task is balanced, so $50\%$ accuracy corresponds to random guessing. Lower detection accuracy indicates stronger resistance to steganalysis.

\subsection{Performance of StyleStegaNet}
\label{sec:exp-stage-performance}

Table~\ref{tab:stage-performance} presents the visual quality of StyleStegaNet at different training stages, reflecting the progressive effects of staged optimization on stego generation, stylized transmission, and secret extraction. The reported values are the DIV2K PSNR thresholds used to determine whether each stage has reached its completion criterion during training.

\begin{table}[!t]
\centering
\caption{DIV2K training-stage completion PSNR of StyleStegaNet.}
\label{tab:stage-performance}
\footnotesize
\setlength{\tabcolsep}{3pt}
\begin{tabular}{llll}
\hline
Stage & Objective & Evaluated pair & PSNR(dB) \\
\hline
Stage 1 & $\mathcal{L}_{\mathrm{S1}}$ & $(\mathbf{X},\mathbf{C})$ & 38.63 \\
Stage 2 & $\mathcal{L}_{\mathrm{S2}}$ & $(\mathbf{X}',\mathbf{X})$ & 32.78 \\
Stage 3 & $\mathcal{L}_{\mathrm{S3}}$ & $(\mathbf{S}',\mathbf{S})$ & 28.49 \\
\hline
\end{tabular}
\end{table}

Table~\ref{tab:stage-performance} shows the DIV2K PSNR values used to determine whether each progressive training stage has reached its completion criterion. Stage~1 reaches $38.63$~dB on $(\mathbf{X},\mathbf{C})$, indicating that the steganographic encoder can embed the secret while keeping the stego image close to the cover. Stage~2 reaches $32.78$~dB on $(\mathbf{X}',\mathbf{X})$, showing that the RevStego network can reconstruct the stego image after stylized transmission. Stage~3 reaches $28.49$~dB on $(\mathbf{S}',\mathbf{S})$, indicating that the full receiver path can extract the secret with sufficient fidelity. These results support the feasibility of the proposed progressive optimization before any comparison with external hiding methods.

\begin{table*}[!t]
\centering
\caption{Stylized-transmission stress test on $(\mathbf{S}',\mathbf{S})$. Conventional baselines use stego $\rightarrow$ stylization $\rightarrow$ original decoder.}
\label{tab:sota-style}
\scriptsize
\resizebox{\textwidth}{!}{%
\begin{tabular}{l|rrrr|rrrr}
\hline
\multicolumn{9}{c}{Stylized transmission: Secret/Recovered image pair $(\mathbf{S}',\mathbf{S})$} \\
\hline
Methods & \multicolumn{4}{c|}{DIV2K} & \multicolumn{4}{c}{MS-COCO} \\
 & PSNR(dB)$\uparrow$ & SSIM$\uparrow$ & APD$\downarrow$ & RMSE$\downarrow$
 & PSNR(dB)$\uparrow$ & SSIM$\uparrow$ & APD$\downarrow$ & RMSE$\downarrow$ \\
\hline
Deep Steganography~\cite{baluja2017hiding} & 7.71 & 0.1106 & 90.37 & 108.36 & 8.07 & 0.1387 & 85.65 & 103.64 \\
HiDDeN~\cite{zhu2018hidden} & 9.76 & 0.1014 & 70.46 & 85.97 & 10.18 & 0.1192 & 67.52 & 82.28 \\
HiNet~\cite{jing2021hinet} & 7.94 & 0.0443 & 85.24 & 103.49 & 8.07 & 0.0415 & 83.83 & 101.94 \\
LSB & 7.83 & 0.0116 & 85.30 & 104.04 & 7.83 & 0.0128 & 85.50 & 104.07 \\
PUSNet~\cite{li2024pusnet} & 7.29 & 0.0286 & 91.75 & 111.21 & 7.82 & 0.0321 & 85.37 & 104.54 \\
WengNet~\cite{weng2019high} & 10.68 & 0.1175 & 65.33 & 75.80 & 10.98 & 0.1409 & 62.89 & 73.66 \\
\textbf{StyleStegaNet} & \textbf{27.02} & \textbf{0.8830} & \textbf{7.71} & \textbf{12.07}
& \textbf{30.37} & \textbf{0.8862} & \textbf{5.41} & \textbf{8.29} \\
\hline
\end{tabular}%
}
\end{table*}

\subsection{Comparison with Existing Methods}
\label{sec:exp-sota}

We compare StyleStegaNet with representative full-image hiding baselines, including Deep Steganography~\cite{baluja2017hiding}, HiDDeN~\cite{zhu2018hidden}, HiNet~\cite{jing2021hinet}, LSB, PUSNet~\cite{li2024pusnet}, and WengNet~\cite{weng2019high}. Table~\ref{tab:sota-style} gives the main stress test under stylized transmission, where conventional baselines follow the pipeline of stego $\rightarrow$ stylization $\rightarrow$ original decoder. The table evaluates the secret/recovered image pair $(\mathbf{S}',\mathbf{S})$.

The quantitative results in Table~\ref{tab:sota-style} demonstrate that applying stylization to the stego image substantially degrades the performance of existing full-image hiding baselines, preventing reliable secret extraction. As can be seen from Table~\ref{tab:sota-style}, when the stego images generated by these methods are directly stylized and then fed into their original decoders, the PSNR of the extracted secret image drops to $7$--$11$~dB. StyleStegaNet significantly outperforms the competing methods across all four secret-extraction metrics. In terms of PSNR, StyleStegaNet achieves $27.02$~dB and $30.37$~dB on DIV2K and MS-COCO, respectively. In addition to PSNR, clear improvements are also observed in SSIM, APD, and RMSE. These gains are mainly attributed to the progressive training strategy and the RevStego network. The former stabilizes the optimization of the complex end-to-end pipeline, while the latter enhances the receiver's ability to extract secret-relevant features from stylized stego images.

\begin{figure}[!t]
\centering
\includegraphics[width=\linewidth]{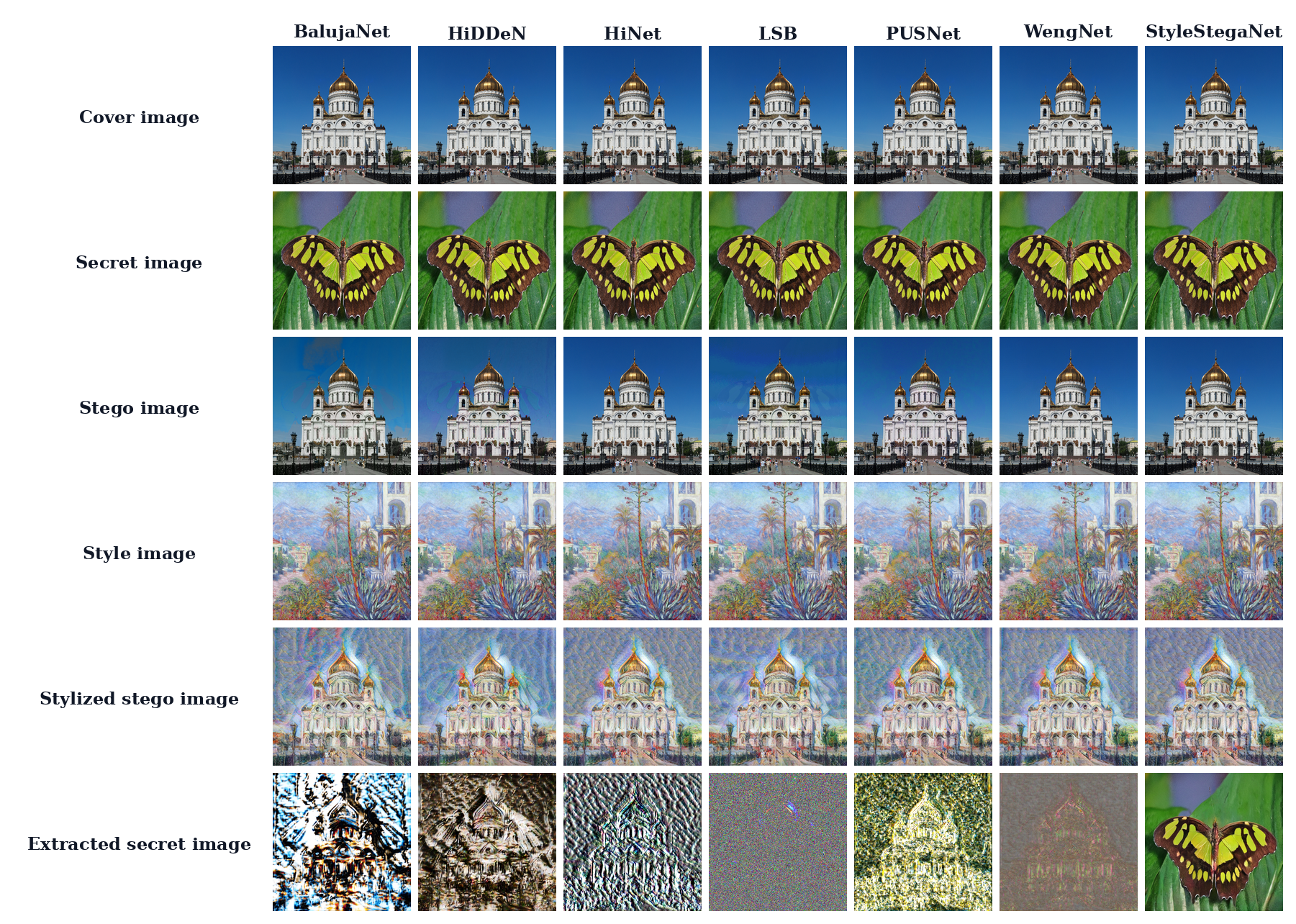}
\caption{Qualitative comparison under stylized transmission. Conventional reveal paths degrade after stylization, while StyleStegaNet remains recoverable.}
\label{fig:stylized-comparison}
\end{figure}

The qualitative results in Figure.~\ref{fig:stylized-comparison} compare the stego, stylized, and extracted secret images of StyleStegaNet and other full-image hiding methods. Although other methods can generate stylized stego images with normal visual appearance, their encoders are not optimized to preserve recoverable stego structures, and post-stylization reconstruction is absent from their training objectives. The results indicate that visual plausibility after stylization is insufficient to ensure secret recoverability. Compared with existing full-image hiding methods, StyleStegaNet achieves a better balance between stylized visual appearance and secret extraction, producing visually plausible stylized stego images while maintaining substantially higher extracted quality of the secret image.

\subsection{Security Analysis}
\label{sec:exp-steganalysis}

\begin{figure}[!t]
\centering
\includegraphics[width=\linewidth]{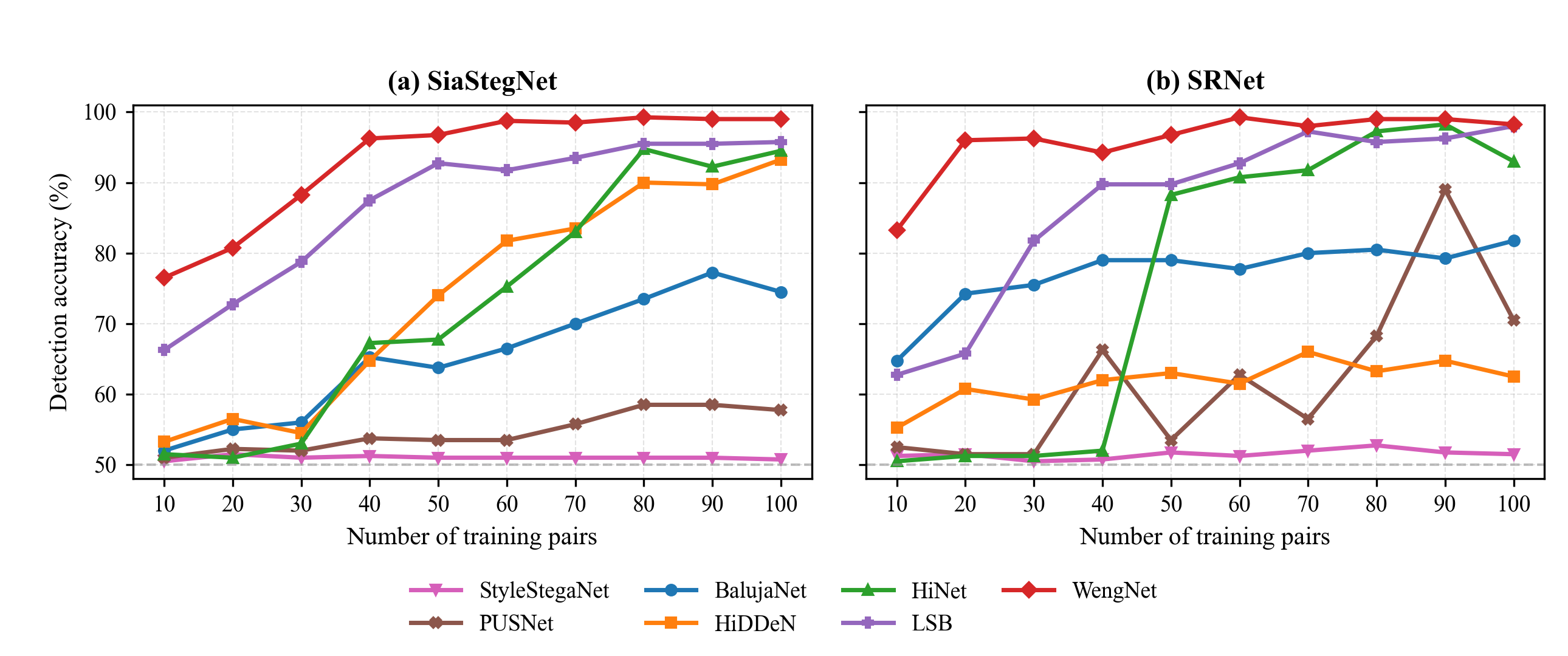}
\caption{Few-shot steganalysis accuracy with the SiaStegNet~\cite{you2021siastegnet} and SRNet~\cite{boroumand2018srnet} detectors. The dashed line marks random guessing; lower accuracy indicates lower detectability.}
\label{fig:steganalysis}
\end{figure}

Security is a fundamental requirement of image steganography for achieving covert communication. In this section, we evaluate the steganalysis resistance of StyleStegaNet and other methods using SRNet~\cite{boroumand2018srnet} and SiaStegNet~\cite{you2021siastegnet}. For each method, the closer the detection accuracy is to $50\%$, the stronger the steganalysis resistance of the image steganography, since $50\%$ accuracy in a binary classification task corresponds to random guessing. The experimental results are reported in Table~\ref{tab:steganalysis} and Figure.~\ref{fig:steganalysis}.

\begin{table}[!b]
\centering
\caption{Deep steganalysis accuracy (\%) of the SiaStegNet~\cite{you2021siastegnet} and SRNet~\cite{boroumand2018srnet} detectors. Lower is better; $50\%$ is random guessing.}
\label{tab:steganalysis}
\scriptsize
\setlength{\tabcolsep}{3pt}
\begin{tabular}{lcccc}
\hline
Method & \multicolumn{2}{c}{SiaStegNet} & \multicolumn{2}{c}{SRNet} \\
       & Avg. & $N{=}100$ & Avg. & $N{=}100$ \\
\hline
\textbf{StyleStegaNet} & \textbf{51.00} & \textbf{50.75} & \textbf{51.50} & \textbf{51.50} \\
PUSNet & 54.65 & 57.75 & 62.23 & 70.50 \\
BalujaNet & 65.38 & 74.50 & 77.17 & 81.75 \\
HiDDeN & 74.12 & 93.25 & 61.83 & 62.50 \\
HiNet & 73.03 & 94.50 & 76.42 & 93.00 \\
LSB & 87.00 & 95.75 & 86.97 & 98.00 \\
WengNet & 93.30 & 99.00 & 96.00 & 98.25 \\
\hline
\end{tabular}
\end{table}

It can be seen from Table~\ref{tab:steganalysis} that StyleStegaNet achieves the lowest average detection accuracy among all evaluated hiding methods for both steganalysis detectors, achieving $51.00\%$ with SiaStegNet and $51.50\%$ with SRNet. The detection accuracy of StyleStegaNet is close to $50\%$, i.e., near random guessing, whereas most existing full-image hiding methods are detected with accuracies higher than $60\%$. This indicates that stylized transmission can conceal the abnormal traces caused by secret embedding at both visual and statistical levels, making the steganographic behavior more difficult to detect.

Figure.~\ref{fig:steganalysis} provides a more intuitive illustration of the trend in detection accuracy as the number of training pairs increases. It can be observed that the detection accuracy of existing full-image hiding methods gradually rises with more training pairs, while StyleStegaNet remains close to $50\%$ with only small variations. This further demonstrates the stronger steganalysis resistance and security of StyleStegaNet.

\subsection{Secret Information Leakage}
\label{sec:exp-leakage}

\begin{figure}[!t]
\centering
\includegraphics[width=\linewidth]{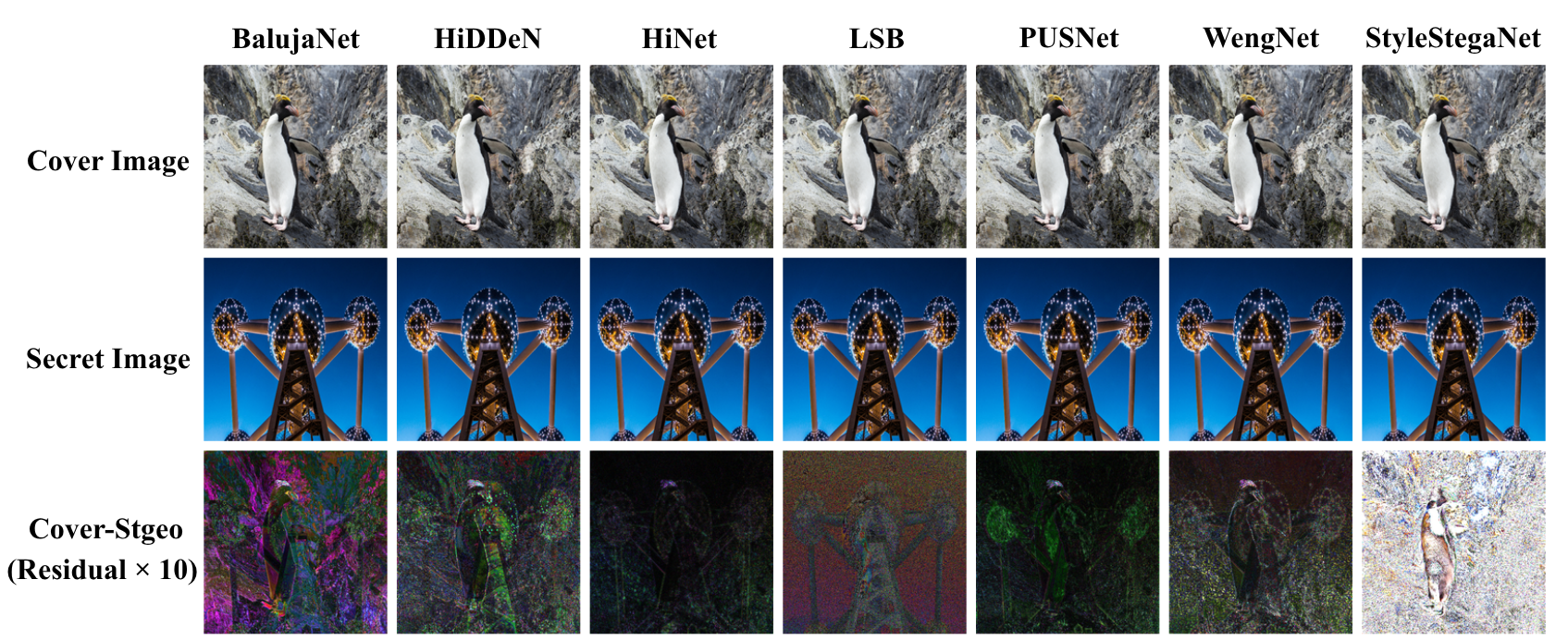}
\caption{Magnified residual maps under stylized transmission ($10{\times}$). In this cover-known diagnostic, StyleStegaNet residuals do not visually expose the secret structure; this is not a guarantee against adaptive attacks.}
\label{fig:residual-comparison}
\end{figure}

Secret information leakage is a critical security concern in existing full-image hiding methods, especially when the original cover image is publicly available or can be obtained by the adversary. Under such a reference-based scenario, the adversary can compare the stego image with the original cover and amplify the residual difference, potentially revealing secret-dependent visual structures. StyleStegaNet is designed to mitigate this leakage by replacing cover-like transmission with stylized transmission. The corresponding qualitative results are presented in Figure.~\ref{fig:residual-comparison}.

Figure.~\ref{fig:residual-comparison} compares the cover--stego residual maps of StyleStegaNet and other full-image hiding methods. In our method, even after the residual between the stego image and the original cover image is amplified by $10{\times}$, no recognizable secret structural information is observed in the residual map. In contrast, for other full-image hiding methods, visually recognizable features of the secret image are exposed in the cover--stego residual images. Although pre-encrypting the secret image can reduce secret information leakage in the residual map, it does not change the fact that existing methods still follow a cover-like transmission paradigm, where hidden communication may be exposed through direct cover--stego comparison. StyleStegaNet breaks the direct correspondence between the stego image and the cover image and camouflages the steganographic behavior at both visual and statistical levels. As a result, it effectively prevents secret information leakage while still enabling high-quality secret extraction, as shown in Figure.~\ref{fig:stylized-comparison}, Row~6.

\subsection{Ablation Study}
\label{sec:exp-ablation}

We perform ablation studies on DIV2K and MS-COCO to verify the effectiveness of progressive three-stage training design. Each ablated variant corresponds to a targeted hypothesis regarding stylized image hiding, and the corresponding results are reported in Table~\ref{tab:ablation}.

\begin{table}[H]
\centering
\caption{Stage~3 ablation on $(\mathbf{S}',\mathbf{S})$. Arrows indicate the preferred direction.}
\label{tab:ablation}
\scriptsize
\setlength{\tabcolsep}{3pt}
\renewcommand{\arraystretch}{0.92}
\begin{tabular}{lrrrr}
\hline
\multicolumn{5}{c}{DIV2K: $(\mathbf{S}',\mathbf{S})$} \\
\hline
Methods & PSNR$\uparrow$ & SSIM$\uparrow$ & APD$\downarrow$ & RMSE$\downarrow$ \\
\hline
\textbf{Full StyleStegaNet} & \textbf{27.02} & \textbf{0.8830} & \textbf{7.71} & \textbf{12.07} \\
Context refiner & 24.91 & 0.8144 & 9.90 & 15.35 \\
w/o secret refinement & 23.83 & 0.7663 & 11.42 & 17.29 \\
w/o high-frequency loss & 24.86 & 0.8065 & 9.96 & 15.44 \\
Late secret refinement & 24.72 & 0.8056 & 10.14 & 15.69 \\
\hline
\multicolumn{5}{c}{MS-COCO: $(\mathbf{S}',\mathbf{S})$} \\
\hline
\textbf{Full StyleStegaNet} & \textbf{30.37} & \textbf{0.8862} & \textbf{5.41} & \textbf{8.29} \\
Context refiner & 29.25 & 0.8689 & 6.21 & 9.42 \\
w/o secret refinement & 28.03 & 0.8382 & 7.13 & 10.75 \\
w/o high-frequency loss & 29.13 & 0.8639 & 6.27 & 9.53 \\
Late secret refinement & 29.01 & 0.8635 & 6.36 & 9.67 \\
\hline
\end{tabular}
\end{table}

As can be seen from Table~\ref{tab:ablation}, the context-refiner variant underperforms the full model on both datasets, indicating that recovery-side context does not bring clear benefits to secret extraction in this setting. Removing the secret refinement branch reduces the PSNR of extracted secret image by $3.18$~dB on DIV2K and $2.34$~dB on MS-COCO, demonstrating that the secret refinement branch is crucial for high-fidelity secret reconstruction. Removing high-frequency loss leads to consistent performance degradation, with the PSNR of extracted secret image reduced by $2.16$~dB and $1.24$~dB on both datasets, respectively. This confirms the importance of explicit frequency-domain constraints in preserving fine structures and texture details that are susceptible to stylization. Late secret refinement performs worse than joint refinement, indicating that sender-receiver alignment during the final training stage is more effective than post-hoc correction.

\FloatBarrier

\section{Discussion and Future Work}
\label{sec:discussion}

 Why stylized transmission matters. Existing full-image hiding methods are largely built on a cover-like assumption, where the stego image is expected to be visually close to the original cover. While this assumption benefits visual fidelity, it also introduces a security vulnerability: any exposure or accurate estimation of the cover provides a clean reference for detecting steganographic behavior. The central aspect of StyleStegaNet is its reformulation of the transmission objective: the transmitted image is no longer constrained to be a pixel-level approximation of the cover, but is required to form a visually coherent stylized stego image consistent with the reference style. The receiver can still extract the secret through structure reconstruction and secret decoding, whereas the adversary can no longer rely on direct cover--stego comparison as primary evidence of steganographic behavior.

 Security interpretation. The experimental results for steganalysis show that, under the few-shot setting, SRNet and SiaStegNet detect StyleStegaNet at accuracies close to random guessing. This observation agrees with the detectability bounds in Propositions~\ref{prop:detect} and~\ref{prop:embed-detect}: embedding the secret into stylization-resilient structural features under a frozen stylization channel limits the induced distributional discrepancy, thereby reducing distinguishability under the evaluated steganalysis detectors.

 Limitation and Future Work. Despite these promising results, StyleStegaNet has two limitations: First, the current formulation assumes that the receiver knows the stylization family and has access to the style reference used by the sender. Second, the structure reconstruction backbone dominates the training cost, suggesting that lighter restoration architectures could improve deployment efficiency. Recent state-space models used in image restoration and steganography, such as MambaIR~\cite{guo2024mambair} and StegMamba~\cite{luo2025stegmamba}, provide possible directions for lightweight receiver design. In future work, we will examine cross-family style shifts and additional channel distortions, including JPEG compression, Gaussian noise, blur, and resizing. We will also explore pre-encryption or key-whitening of the secret image for high-security applications, so that any residual leakage is not directly associated with semantic secret content.

\bibliographystyle{IEEEtran}
\bibliography{ref}

\end{document}